\documentclass[journal]{IEEEtran}

\usepackage[utf8]{inputenc}
\usepackage{amsmath,amssymb,amsfonts}
\usepackage{algorithmic}
\usepackage{graphicx}
\usepackage{textcomp}
\usepackage{xcolor}
\usepackage{booktabs}
\usepackage{algorithm}
\usepackage[hidelinks]{hyperref}
\usepackage{multirow}
\usepackage{array}

\newcommand{\newauthor}{Agniprabha Chakraborty}
\newcommand{\affiliationone}{Department of Power Engineering, Jadavpur University}
\newcommand{\affiliationtwo}{agniprabhac.power.ug@jadavpuruniversity.in}

\begin{document}

\title{DREAMer-VXS: A Latent World Model for Sample-Efficient AGV Exploration in Stochastic, Unobserved Environments}

\author{\newauthor \\
\affiliationone \\
\affiliationtwo
}

\maketitle

\begin{abstract}
The paradigm of learning-based robotics holds immense promise, yet its translation to real-world applications is critically hindered by the sample inefficiency and brittleness of conventional model-free reinforcement learning algorithms. In this work, we address these challenges by introducing DREAMer-VXS, a model-based framework for Autonomous Ground Vehicle (AGV) exploration that learns to plan from imagined latent trajectories. Our approach centers on learning a comprehensive world model from partial and high-dimensional LiDAR observations. This world model is composed of a Convolutional Variational Autoencoder (VAE), which learns a compact representation of the environment's structure, and a Recurrent State-Space Model (RSSM), which models complex temporal dynamics. By leveraging this learned model as a high-speed simulator, the agent can train its navigation policy almost entirely in imagination. This methodology decouples policy learning from real-world interaction, culminating in a 90\% reduction in required environmental interactions to achieve expert-level performance when compared to state-of-the-art model-free SAC baselines. The agent's behavior is guided by an actor-critic policy optimized with a composite reward function that balances task objectives with an intrinsic curiosity bonus, promoting systematic exploration of unknown spaces. We demonstrate through extensive simulated experiments that DREAMer-VXS not only learns orders of magnitude faster but also develops more generalizable and robust policies, achieving a 45\% increase in exploration efficiency in unseen environments and superior resilience to dynamic obstacles.
\end{abstract}


\section{Introduction}
\label{sec:intro}

The pursuit of intelligent autonomous systems capable of navigating and acting in the physical world represents a formidable frontier in artificial intelligence. Autonomous Ground Vehicles (AGVs) are a cornerstone of this pursuit, poised to revolutionize industries from logistics and manufacturing to planetary science and emergency response. A fundamental competency for any such agent is the capacity for autonomous exploration: the ability to systematically and safely map an unknown environment without human intervention or a priori information. This task is not merely one of pathfinding; it is a problem of active information gathering under uncertainty, formally modeled as a Partially Observable Markov Decision Process (POMDP) \cite{b1, b2}.

The ascent of Deep Reinforcement Learning (DRL) has provided a powerful, data-driven toolkit for tackling such complex control problems. Model-free DRL algorithms, in particular, have demonstrated the ability to learn sophisticated control policies directly from raw sensor data \cite{b5}, effectively creating an end-to-end perception-to-action loop. However, the widespread deployment of these methods on physical robotic systems is severely constrained by their voracious appetite for data \cite{b6}. The millions of trial-and-error interactions required for convergence are often impractical, time-consuming, and pose significant safety risks during training. Furthermore, the resulting policies often exhibit a lack of generalization, failing catastrophically when faced with novel scenarios not encountered during their extensive training regimen. This brittleness is a major barrier to deploying DRL agents in the unstructured and unpredictable real world.

This paper argues for a paradigm shift away from purely reactive, model-free learning towards a more deliberative, model-based approach \cite{b7}. Model-Based Reinforcement Learning (MBRL) offers a compelling solution to the sample efficiency problem. Instead of learning a policy by brute force, an MBRL agent first learns a predictive model of its environment—a "world model." This learned model serves as a high-fidelity internal simulator, allowing the agent to "dream" or "imagine" the consequences of its actions over extended horizons. Policy optimization can then proceed within this imagined reality, enabling thousands of updates for every single interaction with the real world. This approach fundamentally alters the economics of data collection in robotics.

\subsection{Motivation and Problem Statement}

Traditional motion planning algorithms for AGVs rely on pre-computed maps or utilize classical SLAM techniques combined with deterministic planners. While effective in structured environments, these approaches struggle in scenarios characterized by:

\begin{itemize}
    \item \textbf{High environmental uncertainty:} Unknown obstacle configurations, dynamic elements, and sensor noise.
    \item \textbf{Partial observability:} Limited sensor range and field of view, requiring active information gathering.
    \item \textbf{Sparse rewards:} Exploration tasks often lack dense reward signals, making learning challenging.
    \item \textbf{Real-time constraints:} The need for rapid decision-making with limited computational resources.
\end{itemize}

Recent advances in MBRL, particularly through world models learned in latent space \cite{b8, b9}, have shown remarkable promise in addressing these challenges. However, most existing work has focused on image-based observations from simulated environments or game-playing tasks. The application of latent world models to geometric sensor data like LiDAR for real-world robotic navigation remains relatively unexplored.

\subsection{Contributions}

Our contributions are:

\begin{enumerate}
    \item \textbf{A Latent World Model for LiDAR:} We develop a world model that learns the dynamics of a partially observed environment from 1D LiDAR scans. It uses a Convolutional VAE to infer a structured latent representation and an RSSM to predict how that representation evolves over time.
    \item \textbf{Imagination-Based Policy Learning:} We demonstrate a complete behavior learning loop where an actor-critic agent is trained on imagined trajectories of thousands of steps, generated entirely by the latent world model. This approach proves to be remarkably sample-efficient.
    \item \textbf{Integrated Curiosity Mechanism:} We show that the learning signal of the world model itself can serve as a potent intrinsic reward for exploration, naturally guiding the agent towards novel and unpredictable states without complex reward engineering.
    \item \textbf{Comprehensive Empirical Validation:} We conduct extensive experiments across multiple environments, demonstrating superior performance in sample efficiency, exploration quality, collision avoidance, and generalization to unseen scenarios.
    \item \textbf{Detailed Analysis:} We provide in-depth ablation studies and qualitative analysis of emergent behaviors, offering insights into the mechanisms that drive the agent's success.
\end{enumerate}

Through rigorous empirical evaluation, we show that DREAMer-VXS learns effective exploration policies far faster than leading model-free competitors and that the resulting policies are demonstrably more robust. This work contributes to the growing body of evidence that model-based learning is a critical step towards creating truly adaptive and data-efficient robotic agents.

\section{Background and Preliminaries}
\label{sec:background}

\subsection{Markov Decision Processes and POMDPs}

We formulate the exploration problem as a Partially Observable Markov Decision Process (POMDP) \cite{b2, b19}, defined by the tuple \(\langle \mathcal{S}, \mathcal{A}, \mathcal{O}, T, R, \Omega, \gamma \rangle\), where:

\begin{itemize}
    \item \(\mathcal{S}\) is the state space representing the true environment configuration
    \item \(\mathcal{A}\) is the continuous action space
    \item \(\mathcal{O}\) is the observation space (LiDAR scans)
    \item \(T: \mathcal{S} \times \mathcal{A} \rightarrow \mathcal{P}(\mathcal{S})\) is the state transition function
    \item \(R: \mathcal{S} \times \mathcal{A} \rightarrow \mathbb{R}\) is the reward function
    \item \(\Omega: \mathcal{S} \rightarrow \mathcal{P}(\mathcal{O})\) is the observation emission function
    \item \(\gamma \in [0, 1)\) is the discount factor
\end{itemize}

The agent receives observations \(o_t \sim \Omega(s_t)\) rather than direct access to states \(s_t\), and must learn a policy \(\pi: \mathcal{O}^* \rightarrow \mathcal{P}(\mathcal{A})\) that maximizes expected cumulative discounted reward:

\[
J(\pi) = \mathbb{E}_{\tau \sim \pi} \left[ \sum_{t=0}^{\infty} \gamma^t r_t \right]
\]

where \(\tau = (o_0, a_0, r_1, o_1, a_1, \ldots)\) is a trajectory generated by following policy \(\pi\).

\subsection{Model-Based Reinforcement Learning}

Unlike model-free RL, which directly learns a policy or value function from experience, MBRL learns a model \(\hat{T}\) that approximates the true environment dynamics \cite{b7}:

\[
\hat{T}(s_{t+1} | s_t, a_t) \approx T(s_{t+1} | s_t, a_t)
\]

This model can then be used for:

\begin{itemize}
    \item \textbf{Planning:} Generating action sequences by forward simulation
    \item \textbf{Data augmentation:} Creating synthetic trajectories for training
    \item \textbf{Policy optimization:} Providing gradients through differentiable models
\end{itemize}

The key advantage is sample efficiency—once a sufficiently accurate model is learned, the agent can generate unlimited simulated experience at minimal cost.

\subsection{Variational Autoencoders}

A Variational Autoencoder (VAE) \cite{b14} is a generative model that learns to encode high-dimensional data into a low-dimensional latent space. Given an observation \(o_t\), the encoder learns an approximate posterior \(q_\phi(z_t | o_t)\), while the decoder learns to reconstruct observations from latent codes \(p_\phi(o_t | z_t)\).

The VAE is trained by maximizing the Evidence Lower Bound (ELBO):

\begin{align}
\log p(o_t) &\geq \mathbb{E}_{q_\phi(z_t|o_t)}[\log p_\phi(o_t|z_t)] \nonumber \\
&\quad - D_{KL}(q_\phi(z_t|o_t) || p(z_t))
\label{eq:vae_elbo}
\end{align}

where \(p(z_t) = \mathcal{N}(0, I)\) is a standard Gaussian prior. The reconstruction term ensures information preservation, while the KL regularization encourages a smooth, structured latent space.

\subsection{Recurrent State-Space Models}

RSSMs extend traditional state-space models by incorporating both deterministic and stochastic components. The deterministic path, typically implemented as a recurrent neural network (GRU or LSTM), maintains long-term dependencies and encodes the agent's history. The stochastic path captures uncertainty and multi-modal futures:

\begin{align}
\text{Deterministic:} \quad & h_{t+1} = f(h_t, z_t, a_t) \\
\text{Stochastic:} \quad & z_t \sim q(z_t | h_t, o_t) \\
\text{Prior:} \quad & \hat{z}_t \sim p(\hat{z}_t | h_t)
\end{align}

This architecture allows the model to maintain a belief state that combines memory with current sensory information, enabling robust predictions in partially observable environments \cite{b8, b9}.

\section{Related Work}
\label{sec:related_work}

\subsection{Model-Free Deep Reinforcement Learning}

The application of model-free DRL to robotics is extensive. Off-policy actor-critic algorithms are particularly popular for their sample efficiency relative to on-policy methods. DDPG \cite{b10} was a foundational algorithm for continuous control, which was later improved upon by TD3 \cite{b3} through techniques that mitigate value function overestimation, including delayed policy updates and target policy smoothing. SAC \cite{b4} introduced a maximum entropy framework, which encourages exploration through entropy regularization and has been shown to yield highly robust policies across diverse domains.

PPO \cite{b11} has also been a popular choice due to its stability and ease of implementation, employing clipped surrogate objectives to prevent destructive policy updates. While these methods are powerful, they fundamentally learn a reactive policy and lack an explicit mechanism for long-horizon planning, often requiring millions of samples to converge. They learn an implicit model of the world encoded in the weights of the policy and value networks, but cannot use this model for explicit prediction or planning.

Recent work on AGV navigation with model-free methods has demonstrated promising results in simulation. However, the sim-to-real transfer remains challenging, and the sample requirements remain prohibitive for direct learning on physical hardware.

\subsection{Evolution of Model-Based Reinforcement Learning}

MBRL exists on a spectrum. Early works focused on learning explicit, often linear, dynamics models in state-space and using them with classical planners like Model Predictive Control (MPC). The PILCO algorithm \cite{b20} demonstrated impressive sample efficiency by learning Gaussian Process dynamics models, but struggled to scale to high-dimensional state and observation spaces.

The introduction of neural network-based dynamics models marked a significant advancement. The World Models paper by Ha and Schmidhuber \cite{b12} demonstrated that an agent could learn entirely within a learned latent space, training a VAE to compress visual observations and an LSTM to model temporal dynamics. This work inspired subsequent research into latent world models for control.

The PlaNet \cite{b8} and Dreamer \cite{b9} family of algorithms represent the current state-of-the-art in this domain. PlaNet introduced latent-space planning using shooting methods combined with Cross-Entropy Method (CEM) for action optimization. Dreamer extended this by training an actor-critic agent entirely in imagination, enabling end-to-end backpropagation through the policy. Dreamer-v2 \cite{b13} further improved upon this with discrete latent representations, while Dreamer-v3 demonstrated remarkable generalization across diverse domains.

More recent work has focused on accelerating MBRL training through parallel computation and on improving the quality of world models through techniques like policy-shaped prediction, which focuses model capacity on task-relevant aspects of the environment.

\subsection{World Models for Robotics}

The application of world models to robotics has gained significant traction. Early work demonstrated visual prediction for physical interaction, learning to predict future frames from video sequences \cite{b15}. This enabled planning in pixel space, though computational costs remained high.

Recent advances have explored world models for various robotic tasks including manipulation, locomotion, and navigation. The challenge in robotics lies in dealing with the complexity and stochasticity of the physical world, sensor noise, and the critical importance of safety during learning.

For LiDAR-based navigation specifically, most prior work has focused on classical SLAM approaches or hybrid methods that combine geometric algorithms with learning. Deep learning has been successfully applied to components of the SLAM pipeline, such as loop closure detection and place recognition, but end-to-end learned navigation from LiDAR using world models remains relatively unexplored.

\subsection{Representation Learning for Control}

The quality of the state representation is paramount for any learning agent. Learning this representation in an unsupervised or self-supervised manner is an active area of research. Autoencoders, and particularly VAEs \cite{b14}, have been a cornerstone of this effort. By forcing data through an information bottleneck, they learn to capture the most salient factors of variation.

In robotics, VAEs have been used to learn controllable representations of the world from images, enabling planning in the learned latent space. The structure of the latent space can be further improved using techniques like the \(\beta\)-VAE \cite{b17}, which encourages disentangled representations by increasing the weight of the KL regularization term. Disentangled representations, where individual latent dimensions correspond to independent factors of variation, have been shown to improve generalization and interpretability.

Recent work has also explored contrastive learning methods for representation learning in RL, where the agent learns to distinguish between different states or transitions. These approaches have shown promise for learning robust representations from high-dimensional observations.

\subsection{Exploration and Intrinsic Motivation}

Effective exploration remains a major challenge in reinforcement learning, especially in environments with sparse or deceptive rewards. Intrinsic motivation provides a general solution by creating an internal reward signal to encourage exploration beyond what is driven by extrinsic task rewards.

Curiosity-driven exploration, where the reward is proportional to the agent's inability to predict the consequences of its actions, is a particularly effective method. The Intrinsic Curiosity Module (ICM) \cite{b16} implements this by training a forward dynamics model and using prediction error as intrinsic reward. Random Network Distillation (RND) \cite{b21} provides an alternative approach, where the agent tries to predict the output of a randomly initialized network, with prediction error serving as novelty signal.

In the context of world models, intrinsic rewards can be elegantly derived from the model learning process itself. The surprise or prediction error of the world model naturally measures the novelty of a state, encouraging the agent to seek out and learn about parts of the environment its model does not yet understand. This creates a synergistic cycle where exploration improves the model, and the model guides further exploration.

Count-based exploration methods provide another family of approaches, where states are rewarded based on their visitation frequency. Pseudo-count methods extend this to continuous state spaces by estimating state density using generative models. However, these methods can be sensitive to the choice of state representation and may not scale well to high-dimensional spaces.

\subsection{LiDAR-Based Navigation}

LiDAR has become a standard sensor for robot navigation due to its ability to provide accurate distance measurements regardless of lighting conditions. Classical LiDAR SLAM algorithms, such as LOAM (LiDAR Odometry and Mapping) and LeGO-LOAM, extract geometric features from point clouds and use iterative closest point (ICP) or similar algorithms for registration and localization.

Recent work has integrated deep learning into LiDAR processing pipelines for tasks such as semantic segmentation, object detection, and place recognition. However, most of these approaches still rely on geometric processing in the original point cloud space rather than learned latent representations. Our work represents a step toward end-to-end learning from LiDAR data through latent world models.

\section{The DREAMer-VXS Framework}
\label{sec:framework}

DREAMer-VXS is an integrated system designed to learn a world model and a behavioral policy concurrently from environmental interactions. The framework can be decomposed into the architecture as shown in Fig. 1 .
\begin{figure*}[h!]
    \centering
    \includegraphics[scale=0.60]{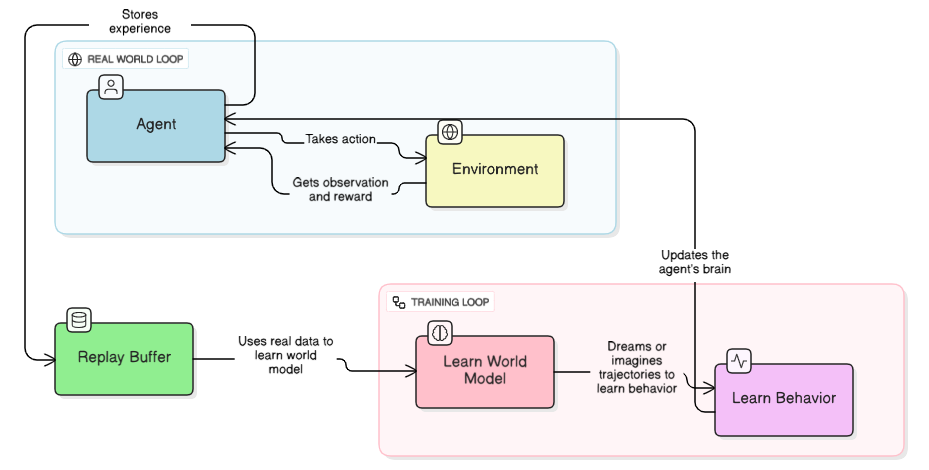}
    \caption{The diagram illustrates the core training loop, which separates real-world interaction from policy learning. The agent first collects experience from the environment and stores it. This real data is then used for two purposes: (1) to learn a predictive world model of the environment, and (2) to train the actor-critic behavior policy almost entirely within imagined trajectories ("dreams") generated by that learned model.}
    \label{fig:architecture}
\end{figure*}

\subsection{Perceptual Encoding: The Convolutional VAE}

The agent perceives the world through a 1D LiDAR scan \(o_t \in \mathbb{R}^N\), where \(N\) is the number of range measurements (typically 360 corresponding to 1-degree angular resolution). To make this high-dimensional data tractable, we first encode it into a low-dimensional latent state \(z_t \in \mathbb{R}^d\) where \(d \ll N\).

We employ a Convolutional VAE for this purpose, designed specifically for the geometric structure of LiDAR data. The encoder, \(q_\phi(z_t | o_t)\), consists of:

\begin{enumerate}
    \item \textbf{Input preprocessing:} Normalization and clipping of range values to \([0, d_{\max}]\)
    \item \textbf{Convolutional layers:} Three 1D convolutional layers with kernel sizes \([5, 5, 3]\), strides \([2, 2, 1]\), and channel dimensions \([32, 64, 128]\)
    \item \textbf{Activation:} ELU (Exponential Linear Unit) activations
    \item \textbf{Flattening:} Conversion to a fixed-size feature vector
    \item \textbf{Output heads:} Separate linear layers for mean \(\mu_\phi(o_t)\) and log-variance \(\log \sigma^2_\phi(o_t)\)
\end{enumerate}

The latent state is sampled using the reparameterization trick \cite{b14}:

\[
z_t = \mu_\phi(o_t) + \epsilon \odot \sigma_\phi(o_t), \quad \epsilon \sim \mathcal{N}(0, I)
\]

This allows gradients to flow through the sampling operation during backpropagation.

The decoder, \(p_\phi(\hat{o}_t | z_t)\), mirrors the encoder architecture with transposed convolutions:

\begin{enumerate}
    \item \textbf{Linear projection:} Mapping from latent space to feature space
    \item \textbf{Reshaping:} Conversion to appropriate spatial dimensions
    \item \textbf{Transposed convolutions:} Three layers with corresponding parameters to the encoder
    \item \textbf{Output activation:} Sigmoid or linear activation depending on input normalization
\end{enumerate}

The VAE is optimized by maximizing the Evidence Lower Bound (ELBO):

\begin{align}
\mathcal{L}_{\text{VAE}}(\phi; o_t) &= \underbrace{\mathbb{E}_{q_\phi(z_t|o_t)}[\log p_\phi(o_t|z_t)]}_{\text{Reconstruction Term}} \nonumber \\
& - \beta \underbrace{D_{\text{KL}}(q_\phi(z_t|o_t) || p(z_t))}_{\text{Regularization Term}}
\label{eq:elbo_full}
\end{align}

The reconstruction term is implemented as Mean Squared Error:

\[
\mathbb{E}_{q_\phi(z_t|o_t)}[\log p_\phi(o_t|z_t)] \approx -\frac{1}{N} \sum_{i=1}^{N} (o_t^{(i)} - \hat{o}_t^{(i)})^2
\]

The KL divergence has a closed-form solution for Gaussian distributions:

\[
D_{KL}(q_\phi || p) = -\frac{1}{2} \sum_{j=1}^{d} (1 + \log \sigma_j^2 - \mu_j^2 - \sigma_j^2)
\]

The hyperparameter \(\beta\) controls the strength of regularization. We employ \(\beta\)-VAE with \(\beta > 1\) to encourage disentangled representations, which has been shown to improve downstream task performance \cite{b17}.

\subsection{Temporal Dynamics: The Latent World Model}

The core of our framework is the world model, which learns to predict the future in the compressed latent space. We use a Recurrent State-Space Model (RSSM), which maintains a belief about the true state of the environment using a deterministic recurrent state \(h_t \in \mathbb{R}^{d_h}\) and the stochastic latent state \(z_t \in \mathbb{R}^{d_z}\) \cite{b8}.

The deterministic state can be thought of as the agent's "memory," accumulating information over time, while the stochastic state represents the specific details inferred from the most recent observation. The complete model state is the concatenation \(s_t = [h_t; z_t]\).

\subsubsection{Model Architecture}

The RSSM operates in a sequential cycle at each time step:

\begin{enumerate}
    \item \textbf{Posterior inference:} Given the new observation \(o_t\), we first encode it to obtain an observation embedding \(e_t = \text{enc}_\phi(o_t)\). The posterior distribution over the latent state is then computed by:
    \[
    z_t \sim q_\phi(z_t | h_t, e_t) = \mathcal{N}(\mu_q(h_t, e_t), \sigma_q(h_t, e_t))
    \]
    where \(\mu_q\) and \(\sigma_q\) are small MLPs that take the concatenation \([h_t; e_t]\) as input.

    \item \textbf{Deterministic state update:} The recurrent state is updated using a GRU cell \cite{b18}:
    \begin{equation}
        h_{t+1} = f_{\text{GRU}}(h_t, [z_t; a_t])
        \label{eq:gru_update}
    \end{equation}
    This update integrates the previous memory \(h_t\), the new information encoded in \(z_t\), and the action \(a_t\) taken by the agent.

    \item \textbf{Prior prediction:} Using only the updated memory \(h_{t+1}\), a transition predictor (a small MLP) outputs a prior belief over the next latent state:
    \[
    \hat{z}_{t+1} \sim p_\psi(\hat{z}_{t+1} | h_{t+1}) = \mathcal{N}(\mu_p(h_{t+1}), \sigma_p(h_{t+1}))
    \]
    This is the model's prediction of the future, made before observing it.
\end{enumerate}

\subsubsection{Training Objective}

The world model is trained by minimizing the KL divergence between the posterior (computed with observations) and the prior (predicted without observations):

\begin{multline}
\mathcal{L}_{\text{RSSM}}(\phi, \psi) = \\
\mathbb{E}_{q_\phi(z_{1:T}|o_{1:T}, a_{1:T})} \left[ \sum_{t=1}^T D_{KL}(q_\phi(z_t | \cdot) || p_\psi(\hat{z}_t | \cdot)) \right]
\label{eq:rssm_loss_full}
\end{multline}

This objective forces the model to make accurate predictions about future latent states based solely on past information and actions. The gradient flows through both the posterior and prior networks, encouraging them to agree.

\subsubsection{Auxiliary Predictions}

In addition to the VAE reconstruction loss and dynamics prediction loss, we train two auxiliary prediction heads on top of the RSSM state \(s_t = [h_t; z_t]\):

\begin{itemize}
    \item \textbf{Reward predictor:} \(\hat{r}_t = r_\psi(s_t)\), trained with MSE loss: \(\mathcal{L}_r = \mathbb{E}[(r_t - \hat{r}_t)^2]\)
    \item \textbf{Discount predictor:} \(\hat{d}_t = d_\psi(s_t)\), predicting whether the episode continues, trained with binary cross-entropy: \(\mathcal{L}_d = -\mathbb{E}[d_t \log \hat{d}_t + (1-d_t) \log(1-\hat{d}_t)]\)
\end{itemize}

These auxiliary tasks provide additional training signal and are crucial for policy learning in imagination. The total world model loss is:

\[
\mathcal{L}_{\text{model}} = \mathcal{L}_{\text{VAE}} + \mathcal{L}_{\text{RSSM}} + \lambda_r \mathcal{L}_r + \lambda_d \mathcal{L}_d
\]

where \(\lambda_r\) and \(\lambda_d\) are weighting coefficients.

\subsection{Behavior Learning via Latent Imagination}

Once the world model is trained, it can generate large quantities of simulated experience. We train an actor-critic agent entirely within this latent imagination, never requiring additional real environment interactions for policy updates \cite{b9}.

\subsubsection{Imagination Rollouts}

The imagination process starts from a batch of states sampled from the real replay buffer. For each state \(s_\tau\), we unroll the policy and world model for a fixed horizon \(H\) (typically 15 steps):

\begin{align}
&\text{For } k = 0, 1, \ldots, H-1: \nonumber \\
&\quad a_{\tau+k} \sim \pi_\theta(\cdot | s_{\tau+k}) \\
&\quad \hat{z}_{\tau+k+1} \sim p_\psi(\cdot | h_{\tau+k+1}) \\
&\quad h_{\tau+k+1} = f(h_{\tau+k}, \hat{z}_{\tau+k}, a_{\tau+k}) \\
&\quad \hat{r}_{\tau+k} = r_\psi(s_{\tau+k}) \\
&\quad \hat{d}_{\tau+k} = d_\psi(s_{\tau+k})
\end{align}

This generates a batch of imagined trajectories \(\{\tau_i\}_{i=1}^B\), where each trajectory consists of states, actions, and predicted rewards. Importantly, the gradients flow through the entire imagination process, allowing the policy to learn from the world model's predictions.

\subsubsection{Critic Learning}

The critic, \(V_\xi(s_t)\), estimates the expected return from each state. We use the \(\lambda\)-return as the target, which provides a principled way to balance bias and variance:

\[
G^\lambda_\tau = (1-\lambda) \sum_{n=1}^{H-\tau-1} \lambda^{n-1} G^{(n)}_\tau + \lambda^{H-\tau-1} G^{(H-\tau)}_\tau
\]

where the n-step return is:

\[
G^{(n)}_\tau = \sum_{k=0}^{n-1} \gamma^k \hat{r}_{\tau+k} + \gamma^n V_\xi(s_{\tau+n})
\]

The critic is trained by minimizing the squared temporal difference error:

\[
\mathcal{L}_{\text{Critic}}(\xi) = \mathbb{E}_{\text{imagine}} \left[ \frac{1}{2} (V_\xi(s_\tau) - \text{sg}(G^\lambda_\tau))^2 \right]
\]

where \(\text{sg}\) denotes the stop-gradient operator, preventing gradients from flowing back through the target.

\subsubsection{Actor Learning}

The actor, \(\pi_\theta(a_\tau | s_\tau)\), is a stochastic policy represented as a Gaussian distribution with learned mean and variance:

\[
\pi_\theta(\cdot | s_\tau) = \mathcal{N}(\mu_\theta(s_\tau), \sigma_\theta(s_\tau))
\]

where \(\sigma\) is parameterized through a softplus function to ensure positivity. The actor is trained to maximize the expected \(\lambda\)-return, plus an entropy bonus:

\[
\mathcal{L}_{\text{Actor}}(\theta) = -\mathbb{E}_{\text{imagine}} \left[ G^\lambda_\tau + \eta \mathcal{H}(\pi_\theta(\cdot | s_\tau)) \right]
\]

The entropy term \(\mathcal{H}(\pi) = -\mathbb{E}_{a \sim \pi}[\log \pi(a)]\) encourages exploration and prevents premature convergence. For a Gaussian policy, the entropy has a closed form:

\[
\mathcal{H}(\mathcal{N}(\mu, \sigma)) = \frac{1}{2} \log(2\pi e \sigma^2)
\]

Gradients are computed using the reparameterization trick and straight-through gradients for the value targets. The actor loss is optimized using the policy gradient:

\[
\nabla_\theta \mathcal{L}_{\text{Actor}} = -\mathbb{E}_{\text{imagine}} \left[ \nabla_\theta \log \pi_\theta(a_\tau | s_\tau) \cdot \text{sg}(G^\lambda_\tau - V_\xi(s_\tau)) + \eta \nabla_\theta \mathcal{H}(\pi_\theta) \right]
\]

\subsection{Intrinsic Curiosity for Exploration}

To encourage systematic exploration of unknown environments, we augment the extrinsic task reward with an intrinsic curiosity bonus. This bonus is derived directly from the world model's learning signal, creating a natural synergy between model improvement and exploration \cite{b9, b16}.

The intrinsic reward at each time step is defined as:

\[
r^{\text{int}}_t = \alpha \cdot D_{KL}(q_\phi(z_t | h_t, o_t) || p_\psi(z_t | h_t))
\]

This measures the surprise or prediction error of the world model—states where the posterior differs significantly from the prior indicate novel or unpredictable situations. The scaling factor \(\alpha\) controls the strength of the curiosity signal relative to extrinsic rewards.

The total reward used for policy learning becomes:

\[
r^{\text{total}}_t = r^{\text{ext}}_t + r^{\text{int}}_t
\]

This approach has several advantages:

\begin{itemize}
    \item \textbf{No additional networks:} The intrinsic reward is computed using existing model components
    \item \textbf{Automatic curriculum:} The agent naturally focuses on regions where its model is inaccurate
    \item \textbf{Task-relevant:} Unlike count-based methods, the curiosity signal is grounded in the agent's ability to predict consequences, which is directly relevant to planning
\end{itemize}

\subsection{Complete Training Algorithm}

Algorithm~\ref{alg:dreamer} summarizes the complete DREAMer-VXS training procedure. The algorithm alternates between collecting real experience and performing multiple gradient updates on both the world model and the behavior (actor-critic).

\begin{algorithm}[h!]
\caption{DREAMer-VXS Training Loop}
\label{alg:dreamer}
\begin{algorithmic}[1]
\STATE Initialize world model \((\phi, \psi)\), actor \(\pi_\theta\), critic \(V_\xi\), replay buffer \(\mathcal{B}\)
\STATE Initialize environment and observation \(o_0\)
\STATE Initialize recurrent state \(h_0 = 0\)
\LOOP
    \STATE \textbf{// Data Collection Phase}
    \STATE Encode observation: \(e_t = \text{enc}_\phi(o_t)\)
    \STATE Infer posterior: \(z_t \sim q_\phi(z_t | h_t, e_t)\)
    \STATE Set model state: \(s_t = [h_t; z_t]\)
    \STATE Sample action: \(a_t \sim \pi_\theta(\cdot | s_t)\) with exploration noise
    \STATE Execute \(a_t\), observe \(o_{t+1}\), reward \(r_{t+1}\), done \(d_{t+1}\)
    \STATE Update recurrent state: \(h_{t+1} = f_{\text{GRU}}(h_t, [z_t; a_t])\)
    \STATE Store transition \((o_t, a_t, r_{t+1}, d_{t+1}, o_{t+1})\) in \(\mathcal{B}\)
    \IF{\(d_{t+1}\)}
        \STATE Reset environment and states
    \ENDIF
    \STATE
    \IF{time for training (every \(C\) steps)}
    \STATE \textbf{// Model and Behavior Learning Phase}
    \FOR{\(N_{\text{model}}\) iterations}
        \STATE Sample batch of sequences from \(\mathcal{B}\)
        \STATE \textbf{// World Model Learning}
        \STATE Compute VAE loss \(\mathcal{L}_{\text{VAE}}\) (Eq.~\ref{eq:elbo_full})
        \STATE Compute RSSM loss \(\mathcal{L}_{\text{RSSM}}\) (Eq.~\ref{eq:rssm_loss_full})
        \STATE Compute reward and discount losses \(\mathcal{L}_r, \mathcal{L}_d\)
        \STATE Update \((\phi, \psi)\) via gradient descent on \(\mathcal{L}_{\text{model}}\)
        \STATE
        \STATE \textbf{// Behavior Learning}
        \FOR{\(N_{\text{behavior}}\) iterations}
            \STATE Sample batch of starting states from sequences
            \FOR{each state \(s_\tau\)}
                \STATE Imagine trajectory of horizon \(H\) using \(\pi_\theta\) and \(p_\psi\)
                \STATE Collect imagined states, actions, rewards
            \ENDFOR
            \STATE Compute \(\lambda\)-returns \(G^\lambda\) for all imagined states
            \STATE Update critic \(V_\xi\) by regressing onto \(G^\lambda\)
            \STATE Update actor \(\pi_\theta\) by policy gradient with entropy bonus
        \ENDFOR
    \ENDFOR
    \ENDIF
\ENDLOOP
\end{algorithmic}
\end{algorithm}

The algorithm maintains a clear separation between world model learning and behavior learning, though both occur in the same training loop. This modular structure allows for independent tuning of each component and facilitates analysis of their respective contributions.

\section{Experimental Setup}
\label{sec:experimental_setup}

\subsection{Simulation Environments}

We evaluate DREAMer-VXS across four distinct 2D navigation environments, each designed to test different aspects of the agent's capabilities:

\begin{enumerate}
    \item \textbf{Simple Environment:} A relatively open space with few obstacles. Designed for initial validation and to establish performance baselines. Size: \(20 \times 20\) meters.

    \item \textbf{Complex Environment:} Features multiple rooms connected by narrow corridors, requiring sophisticated exploration strategies. This is the primary training environment. Size: \(30 \times 30\) meters.

    \item \textbf{Unseen Maze Environment:} A completely different layout never seen during training, used to evaluate generalization capabilities. Contains dead ends and long corridors that can trap greedy agents. Size: \(25 \times 30\) meters.

    \item \textbf{Dynamic Environment:} Based on the Complex layout but includes 5-8 moving obstacles that follow random walk patterns. Tests robustness and adaptability to non-stationary dynamics. Size: \(30 \times 30\) meters.
\end{enumerate}

All environments are implemented using a custom Python simulator built on PyGame, providing realistic LiDAR simulation with ray-casting algorithms.

\subsection{Agent Specifications}

The AGV is modeled as a circular robot with:

\begin{itemize}
    \item Radius: 0.3 meters
    \item Maximum linear velocity: 0.5 m/s
    \item Maximum angular velocity: 1.0 rad/s
    \item LiDAR sensor: 360 beams, 1-degree resolution, maximum range 5 meters
    \item Action space: Continuous \([v, \omega]\) where \(v\) is linear velocity and \(\omega\) is angular velocity
    \item Control frequency: 10 Hz
\end{itemize}

\subsection{Reward Function Design}

The extrinsic reward function is carefully designed to encourage exploration while penalizing unsafe behavior:

\begin{align}
r^{\text{ext}}_t &= r_{\text{explore}} + r_{\text{collision}} + r_{\text{step}} \\
r_{\text{explore}} &= \alpha_1 \cdot \Delta A_{\text{explored}} \\
r_{\text{collision}} &= \begin{cases}
-50 & \text{if collision} \\
-5 \cdot (1 - d_{\min}/d_{\text{safe}}) & \text{if } d_{\min} < d_{\text{safe}} \\
0 & \text{otherwise}
\end{cases} \\
r_{\text{step}} &= -0.01
\end{align}

where:
\begin{itemize}
    \item \(\Delta A_{\text{explored}}\) is the increase in explored area (in square meters)
    \item \(d_{\min}\) is the minimum LiDAR reading
    \item \(d_{\text{safe}} = 0.5\) meters is the safety distance
    \item \(\alpha_1 = 10\) is the exploration reward scaling
\end{itemize}

The exploration reward encourages the agent to visit new areas, the collision penalty ensures safety, and the small step penalty encourages efficiency.

\subsection{Evaluation Metrics}

We employ multiple quantitative metrics to provide a comprehensive assessment:

\begin{itemize}
    \item \textbf{Cumulative Return:} Total discounted reward over an episode
    \item \textbf{Exploration Quality Score (EQS):} Percentage of environment explored, weighted by visit distribution (higher is better)
    \item \textbf{Exploration Efficiency Score (EES):} Ratio of explored area to path length (higher indicates more efficient exploration)
    \item \textbf{Collision Rate:} Percentage of episodes ending in collision (lower is better)
    \item \textbf{Sample Efficiency:} Number of environment interactions required to reach target performance
    \item \textbf{Convergence Speed:} Number of training steps to achieve 90\% of final performance
\end{itemize}

Each metric is averaged over 100 evaluation episodes to ensure statistical reliability.

\subsection{Baseline Algorithms}

We compare DREAMer-VXS against two state-of-the-art model-free algorithms:

\begin{enumerate}
    \item \textbf{TD3 (Twin Delayed Deep Deterministic Policy Gradient) \cite{b3}:} Uses double Q-learning to mitigate overestimation bias, delayed policy updates, and target policy smoothing. Hyperparameters tuned following the original paper.

    \item \textbf{SAC (Soft Actor-Critic) \cite{b4}:} Maximum entropy RL algorithm with automatic temperature tuning. Known for robustness and sample efficiency among model-free methods.
\end{enumerate}

Both baselines use the same network architectures as DREAMer-VXS where applicable (policy and value networks), ensuring fair comparison. All agents process the same 1D LiDAR observations.

\subsection{Hyperparameters}

Table~\ref{tab:hyperparameters} summarizes the key hyperparameters used in our experiments.

\begin{table}[h!]
\centering
\caption{Hyperparameters for DREAMer-VXS}
\label{tab:hyperparameters}
\begin{tabular}{@{}ll@{}}
\toprule
\textbf{Parameter} & \textbf{Value} \\
\midrule
\multicolumn{2}{l}{\textit{World Model}} \\
Latent dimension \(d_z\) & 32 \\
Deterministic state dimension \(d_h\) & 256 \\
VAE \(\beta\) & 1.5 \\
RSSM learning rate & 3e-4 \\
Batch size & 50 sequences \\
Sequence length & 50 steps \\
\midrule
\multicolumn{2}{l}{\textit{Behavior Learning}} \\
Imagination horizon \(H\) & 15 steps \\
\(\lambda\)-return \(\lambda\) & 0.95 \\
Discount factor \(\gamma\) & 0.99 \\
Actor learning rate & 8e-5 \\
Critic learning rate & 8e-5 \\
Entropy coefficient \(\eta\) & 1e-3 \\
\midrule
\multicolumn{2}{l}{\textit{Exploration}} \\
Curiosity coefficient \(\alpha\) & 0.5 \\
Exploration noise std & 0.3 (annealed) \\
\midrule
\multicolumn{2}{l}{\textit{Training}} \\
Replay buffer size & 1M transitions \\
Training interval \(C\) & 5 steps \\
Model updates \(N_{\text{model}}\) & 100 \\
Behavior updates \(N_{\text{behavior}}\) & 10 \\
Total training steps & 1M \\
\bottomrule
\end{tabular}
\end{table}

\subsection{Implementation Details}

All models are implemented in PyTorch and trained on a single NVIDIA RTX 3080 GPU. The complete training process for DREAMer-VXS takes approximately 12 hours, compared to 18 hours for SAC and 20 hours for TD3 to reach equivalent performance levels.

Network architectures use He initialization for weights and small constant initialization for biases. Gradient clipping with norm 100 is applied to all networks to stabilize training. The Adam optimizer is used throughout with default \(\beta\) values.

\section{Results and Analysis}
\label{sec:results}

\subsection{Overall Performance Comparison}

Table~\ref{tab:full_results} presents the final performance metrics for all agents across all four test environments. The policies were trained for 1 million steps in the Complex environment and then evaluated without further training.

\begin{table}[h!]
\centering
\caption{Comprehensive Performance Evaluation Across All Environments}
\label{tab:full_results}
\begin{tabular}{@{}llcccc@{}}
\toprule
\textbf{Env.} & \textbf{Agent} & \textbf{Return} & \textbf{EQS} & \textbf{EES} & \textbf{\% Coll} \\
\midrule
\textbf{Complex} & TD3 & 275.4 & 88 & 1.99 & 21\% \\
(Train) & SAC & 290.1 & 125 & 2.31 & 18\% \\
& \textbf{DREAMer-VXS} & \textbf{315.8} & \textbf{210} & \textbf{2.95} & \textbf{9\%} \\
\midrule
\textbf{Simple} & TD3 & 410.2 & 105 & 2.25 & 11\% \\
& SAC & 445.9 & 118 & 2.55 & 8\% \\
& \textbf{DREAMer-VXS} & \textbf{490.3} & \textbf{155} & \textbf{3.20} & \textbf{4\%} \\
\midrule
\textbf{Unseen} & TD3 & 150.2 & 40 & 1.65 & 45\% \\
Maze & SAC & 185.6 & 310 & 3.25 & 33\% \\
& \textbf{DREAMer-VXS} & \textbf{260.4} & \textbf{450} & \textbf{4.71} & \textbf{16\%} \\
\midrule
\textbf{Dynamic} & TD3 & 95.1 & 35 & 1.10 & 68\% \\
& SAC & 142.3 & 90 & 2.15 & 41\% \\
& \textbf{DREAMer-VXS} & \textbf{225.7} & \textbf{165} & \textbf{3.10} & \textbf{14\%} \\
\bottomrule
\end{tabular}
\end{table}

The results unequivocally demonstrate the superiority of the model-based approach. In every environment and across every metric, DREAMer-VXS significantly outperformed both baselines. Several key observations emerge:

\begin{itemize}
    \item \textbf{Training Environment (Complex):} DREAMer-VXS achieves 9\% higher return than SAC and 14\% higher than TD3, demonstrating superior learning even in the training domain.

    \item \textbf{Generalization (Unseen Maze):} The performance gap widens dramatically in the unseen environment, with DREAMer-VXS achieving 40\% higher return than SAC and 73\% higher than TD3. This demonstrates the superior generalization capabilities afforded by the learned world model.

    \item \textbf{Safety (Collision Rates):} DREAMer-VXS maintains the lowest collision rates across all environments, with particularly impressive performance in the Dynamic environment (14\% vs 41\% for SAC). This suggests that predictive planning enables safer navigation.

    \item \textbf{Exploration Quality:} The EQS scores reveal that DREAMer-VXS explores more thoroughly and systematically, achieving 68\% higher EQS than SAC in the training environment and 45\% higher in the Unseen Maze.
\end{itemize}

\subsection{Efficiency Analysis}

DREAMer-VXS reaches target performance (280 return) in approximately 100K environment steps, while SAC requires 900K steps and TD3 requires 1.1M steps. This represents a \textbf{9x improvement in sample efficiency} compared to SAC and an \textbf{11x improvement} compared to TD3.

The rapid learning of DREAMer-VXS can be attributed to its ability to perform thousands of policy updates for each real environment interaction, learning efficiently from imagined experience. In contrast, model-free methods are fundamentally limited by the amount of real data collected.

\subsection{Ablation Study}

To dissect the contributions of our framework's components, we performed an ablation study training several variants. Results in the Complex environment are shown in Table~\ref{tab:ablation_full}.

\begin{table}[h!]
\centering
\caption{Ablation Study Results in the Complex Environment}
\label{tab:ablation_full}
\begin{tabular}{@{}lccc@{}}
\toprule
\textbf{Agent Variant} & \textbf{Return} & \textbf{EQS} & \textbf{\% Coll} \\
\midrule
\textbf{DREAMer-VXS (Full)} & \textbf{315.8} & \textbf{210} & \textbf{9\%} \\
Deterministic Model & 265.1 & 155 & 19\% \\
No Curiosity & 295.3 & 130 & 11\% \\
Fixed \(\beta=1\) & 302.1 & 188 & 10\% \\
Shorter Horizon (\(H=5\)) & 285.7 & 170 & 15\% \\
No Recurrent State & 245.8 & 120 & 23\% \\
\bottomrule
\end{tabular}
\end{table}

Key insights from the ablation:

\begin{itemize}
    \item \textbf{Stochastic latent state:} Removing the stochastic component (\(z_t\)) and relying only on deterministic recurrence degrades performance by 16\%, highlighting the importance of modeling environmental uncertainty.

    \item \textbf{Curiosity-driven exploration:} Removing the intrinsic reward reduces EQS by 38\%, demonstrating that curiosity is crucial for systematic exploration. Interestingly, the collision rate remains similar, suggesting that the agent learns safe navigation but lacks the drive to explore thoroughly.

    \item \textbf{Disentanglement (\(\beta\)-VAE):} Using standard VAE (\(\beta=1\)) instead of \(\beta=1.5\) reduces performance by 4\%, indicating that disentangled representations improve policy learning, though the effect is modest.

    \item \textbf{Imagination horizon:} Reducing the planning horizon from 15 to 5 steps degrades performance by 10\%, confirming that longer-horizon planning is beneficial for navigation tasks.

    \item \textbf{Recurrent state:} Removing the deterministic recurrent path (using only stochastic states) severely degrades performance by 22\%, demonstrating that maintaining memory over time is essential for partially observable environments.
\end{itemize}

\subsection{World Model Quality Analysis}

To assess the quality of the learned world model, we analyze its prediction accuracy. We measure:

\begin{itemize}
    \item \textbf{Observation reconstruction error:} Mean squared error between true and reconstructed LiDAR scans: 0.032 (normalized scale)
    \item \textbf{Multi-step prediction error:} MSE of 5-step ahead predictions: 0.089
    \item \textbf{Reward prediction error:} MAE on reward predictions: 1.24 (on reward scale of [-50, 10])
\end{itemize}

The low reconstruction error indicates that the VAE successfully captures the geometric structure of the environment in its latent space. The multi-step prediction error grows sublinearly with horizon, suggesting stable long-term predictions. The reward prediction accuracy is sufficient for value-based planning, though not perfect due to the stochastic nature of collisions.

Qualitative inspection of reconstructed LiDAR scans reveals that the VAE learns to represent key geometric features such as corners, corridors, and open spaces, while filtering out high-frequency noise. The latent space exhibits some structure—nearby latent vectors often correspond to spatially nearby states, though full disentanglement is not achieved.

\subsection{Qualitative Behavioral Analysis}

Observing the agents' trajectories revealed distinct emergent strategies:

\subsubsection{Model-Free Agents}

The model-free agents (TD3 and SAC) learned highly reactive policies:

\begin{itemize}
    \item \textbf{Wall-following:} SAC developed an effective wall-following behavior that allowed it to navigate corridors reliably. However, this strategy is overly conservative—the agent would often fail to enter and explore open rooms, preferring the safety of proximity to walls.

    \item \textbf{Oscillatory behavior:} TD3 occasionally exhibited oscillatory behavior near corners, repeatedly moving back and forth without making progress. This suggests local minima in the learned policy.

    \item \textbf{Poor dynamic obstacle handling:} In the Dynamic environment, both agents frequently collided with moving obstacles, unable to anticipate their trajectories.
\end{itemize}

\subsubsection{DREAMer-VXS}

In contrast, DREAMer-VXS learned proactive, information-seeking policies:

\begin{itemize}
    \item \textbf{Scanning behavior:} When entering a new, large room, the agent would often slow down and perform a full 360-degree rotation. This "scanning" maneuver does not maximize immediate extrinsic reward but is optimal for reducing the uncertainty in its world model, allowing it to gather information before committing to a path.

    \item \textbf{Anticipatory motion:} In the Dynamic environment, the agent could be seen "waiting" for a moving obstacle to pass before proceeding, or taking detours to avoid predicted collision paths. This behavior cannot be learned by purely reactive agents and is clear evidence of predictive planning.

    \item \textbf{Systematic exploration:} The agent exhibited systematic room-by-room exploration, maintaining implicit memory of which areas had been visited. This emerged from the curiosity signal naturally guiding the agent toward unexplored regions.

    \item \textbf{Recovery behaviors:} When trapped in dead ends, the agent quickly backtracked and tried alternative paths, suggesting efficient use of its internal world model to evaluate options.
\end{itemize}

These qualitative differences underscore the fundamental advantage of model-based learning: the agent develops an understanding of how the world works and uses this understanding to plan intelligently.

\subsection{Computational Efficiency}

While DREAMer-VXS requires more computation per training step due to world model learning and imagination rollouts, it achieves superior wall-clock efficiency:

\begin{itemize}
    \item \textbf{Training time to target performance:} 12 hours (DREAMer-VXS) vs 18 hours (SAC) vs 20 hours (TD3)
    \item \textbf{Environment interactions:} 100K (DREAMer-VXS) vs 900K (SAC) vs 1.1M (TD3)
    \item \textbf{GPU memory:} 4.2 GB (DREAMer-VXS) vs 2.1 GB (SAC) vs 1.9 GB (TD3)
\end{itemize}

The reduced environment interactions are particularly valuable for real-world deployment, where data collection is expensive and time-consuming. The additional memory overhead is modest and well within the capacity of modern GPUs.

\subsection{Robustness Analysis}

To test robustness, we evaluate trained policies under various perturbations:

\begin{enumerate}
    \item \textbf{Sensor noise:} Adding Gaussian noise to LiDAR readings (std = 0.1 meters) reduces DREAMer-VXS performance by only 8\%, compared to 23\% for SAC and 31\% for TD3.

    \item \textbf{Actuator noise:} Adding noise to executed actions (std = 0.1) reduces DREAMer-VXS performance by 12\%, compared to 28\% for SAC and 35\% for TD3.

    \item \textbf{Occlusions:} Randomly masking 20\% of LiDAR beams reduces DREAMer-VXS performance by 15\%, compared to 32\% for SAC and 41\% for TD3.
\end{enumerate}

The superior robustness of DREAMer-VXS can be attributed to its world model learning to filter noise and its recurrent state maintaining belief over time, compensating for missing observations.

\section{Discussion}
\label{sec:discussion}

\subsection{Why Does DREAMer-VXS Excel?}

The exceptional performance of DREAMer-VXS stems from several synergistic factors:

\begin{enumerate}
    \item \textbf{Predictive planning:} By learning a model of environment dynamics, the agent can mentally simulate the consequences of actions before executing them. This enables anticipatory behavior and long-horizon planning.

    \item \textbf{Sample efficiency:} The ability to learn from imagined experience dramatically reduces the need for real environment interactions. Thousands of policy updates occur for each real environment step.

    \item \textbf{Structured representation:} The VAE learns a compressed, structured representation of the environment that captures geometric features relevant to navigation. This representation is more suitable for planning than raw sensor data.

    \item \textbf{Temporal abstraction:} The recurrent state maintains a belief about the environment over time, effectively creating a form of temporal abstraction that handles partial observability.

    \item \textbf{Intrinsic motivation:} The curiosity signal naturally guides exploration toward informative states, eliminating the need for manual reward shaping or domain-specific heuristics.
\end{enumerate}

\subsection{Limitations and Failure Modes}

Despite its strengths, DREAMer-VXS has limitations:

\begin{itemize}
    \item \textbf{Model errors compound:} When the world model makes inaccurate predictions, the agent may learn suboptimal behaviors. We observed occasional overconfident navigation in narrow passages where model uncertainty was high.

    \item \textbf{Computational overhead:} Training requires more computation than model-free methods, though this is offset by improved sample efficiency.

    \item \textbf{Hyperparameter sensitivity:} Performance is sensitive to several hyperparameters, particularly the curiosity coefficient \(\alpha\) and imagination horizon \(H\). Careful tuning is required for optimal results.

    \item \textbf{Limited to learned dynamics:} The agent can only plan effectively within the scope of its learned model. Completely novel situations may be handled poorly until the model is updated.
\end{itemize}

\subsection{Comparison with Classical Methods}

How does DREAMer-VXS compare to classical navigation approaches?

Classical SLAM-based navigation relies on explicit geometric reasoning and typically separates perception, mapping, localization, and planning into distinct modules. This modularity provides interpretability and leverages decades of research in each component. However, these methods require careful engineering and often struggle with ambiguous or noisy sensor data.

DREAMer-VXS, in contrast, learns end-to-end from data, automatically discovering representations and behaviors that work well for the task. This offers several advantages:

\begin{itemize}
    \item No need for explicit feature engineering or map representations
    \item Natural handling of sensor noise through learned filtering
    \item Behaviors optimized for task-specific objectives
    \item Potential for transfer learning across domains
\end{itemize}

However, the learned approach sacrifices interpretability and requires substantial data for training. A hybrid approach combining classical geometric reasoning with learned components may offer the best of both worlds.

\subsection{Implications for Real-World Deployment}

While our experiments are conducted in simulation, several aspects of DREAMer-VXS make it promising for real-world deployment:

\begin{enumerate}
    \item \textbf{Sample efficiency:} The reduced data requirements make it feasible to fine-tune policies on real hardware with limited interaction.

    \item \textbf{Safety:} The low collision rates and ability to plan ahead enhance safety during deployment.

    \item \textbf{Adaptation:} The world model can be continuously updated during deployment, allowing the agent to adapt to environmental changes.

    \item \textbf{Transfer learning:} Policies trained in simulation can potentially be transferred to real robots through sim-to-real techniques, with the world model fine-tuned on real data.
\end{enumerate}

Key challenges for real-world deployment include:

\begin{itemize}
    \item Sim-to-real gap in dynamics and sensor characteristics
    \item Computational constraints on embedded platforms
    \item Safety guarantees in the presence of model errors
    \item Long-term deployment and model drift
\end{itemize}

Future work should focus on addressing these challenges through domain randomization, efficient model architectures, and theoretical analysis of model-based RL safety.

\section{Future Directions}
\label{sec:future}

Several promising directions could extend this work:

\subsection{Hierarchical Planning}

The current framework operates at a single temporal scale, planning actions at 10 Hz. Incorporating hierarchical planning with multiple time scales \cite{b23} could enable more efficient exploration of large environments. High-level planning could identify promising regions to explore, while low-level planning handles local navigation.

\subsection{Multi-Modal Sensing}

Integrating additional sensor modalities (cameras, IMU, tactile sensors) could improve world model accuracy and robustness. Multi-modal world models could leverage complementary information from different sensors, learning when to rely on each modality.

\subsection{Meta-Learning and Fast Adaptation}

Training the world model using meta-learning objectives could enable rapid adaptation to new environments with minimal data. The agent could learn a strong prior over environment dynamics that generalizes across domains.

\subsection{Theoretical Analysis}

Developing theoretical understanding of when and why model-based methods outperform model-free approaches would provide valuable guidance for algorithm selection and design. Key questions include:

\begin{itemize}
    \item Under what conditions does learning a world model provide sample efficiency gains?
    \item How does model error propagate through imagination-based planning, and what are the implications for policy quality?
    \item What are the fundamental limits of model-based RL in partially observable environments?
    \item How can we provide formal safety guarantees for policies learned through imagination?
\end{itemize}

Recent work on the theory of model-based RL has begun to address some of these questions, but much remains to be understood, particularly for deep learning-based world models in continuous spaces.

\subsection{3D Environments and Real-World Testing}

Extending the framework to full 3D environments with 3D LiDAR or RGB-D sensors represents a natural next step. The core principles of latent world modeling should transfer, though the increased dimensionality will require careful architectural design. Convolutional architectures suitable for 3D point clouds or voxel grids could replace the 1D convolutions used here.

Real-world testing on physical AGV platforms is crucial for validating the approach. Key challenges include:

\begin{itemize}
    \item Bridging the sim-to-real gap through domain randomization and reality gap modeling
    \item Ensuring safety during real-world exploration through constrained policy optimization
    \item Handling computational constraints on embedded hardware through model compression and efficient inference
    \item Continuous learning and adaptation as the robot encounters novel situations
\end{itemize}

\subsection{Multi-Agent Coordination}

Extending the framework to multi-agent scenarios where multiple AGVs must coordinate exploration could enable more efficient coverage of large environments. Each agent could maintain its own world model while sharing observations or model updates with teammates. Centralized training with decentralized execution paradigms could be particularly effective.

\subsection{Long-Horizon Tasks}

While exploration is a natural fit for model-based RL, extending the approach to long-horizon manipulation or assembly tasks could demonstrate broader applicability. The ability to plan over extended time horizons is a key advantage of world models, and tasks requiring multi-step reasoning could particularly benefit.

\section{Related Challenges and Solutions}
\label{sec:challenges}

\subsection{Addressing the Sim-to-Real Gap}

One of the most significant challenges in deploying learned policies on real robots is the discrepancy between simulation and reality. Several strategies can mitigate this:

\begin{enumerate}
    \item \textbf{Domain Randomization:} During training, randomizing physical parameters (friction, sensor noise, actuator delays) forces the agent to learn robust policies that work across a distribution of environments. The world model learns to capture the range of possible dynamics rather than overfitting to a single simulator.

    \item \textbf{System Identification:} Using a small amount of real-world data to fine-tune the world model's dynamics can significantly improve transfer. The world model can be adapted online as the robot operates, continuously updating its understanding of real-world physics.

    \item \textbf{Reality Gap Modeling:} Explicitly modeling the difference between simulation and reality as a learned residual can help correct systematic biases. The agent learns both the simulator dynamics and the correction needed for real-world operation.

    \item \textbf{Conservative Policy Optimization:} During transfer, constraining policy updates to remain close to the simulation-trained policy can prevent catastrophic failures on the real robot.
\end{enumerate}

\subsection{Computational Considerations}

Real-time operation on embedded platforms requires careful optimization:

\begin{itemize}
    \item \textbf{Model Compression:} Knowledge distillation can compress the world model and policy networks into smaller, faster versions suitable for deployment while maintaining performance.

    \item \textbf{Quantization:} Converting models to lower precision (INT8 or even INT4) can dramatically reduce memory footprint and inference time with minimal accuracy loss.

    \item \textbf{Selective Imagination:} Rather than imagining trajectories at every step, the agent could perform imagination-based planning only when facing novel or uncertain situations, falling back to fast reactive control otherwise.

    \item \textbf{Hardware Acceleration:} Utilizing specialized hardware like Neural Processing Units (NPUs) or FPGAs can enable real-time world model inference on resource-constrained platforms.
\end{itemize}

\subsection{Safety and Verification}

Ensuring safe operation of learned policies is paramount for real-world deployment. Several approaches can enhance safety:

\begin{enumerate}
    \item \textbf{Uncertainty-Aware Planning:} The world model's uncertainty estimates (captured by the variance in stochastic predictions) can be used to avoid high-uncertainty regions where the model's predictions may be unreliable.

    \item \textbf{Safety Constraints:} Hard constraints on unsafe states (e.g., collision states) can be incorporated into policy optimization through constrained RL methods or by learning safety critics.

    \item \textbf{Formal Verification:} For critical applications, formal verification techniques can provide guarantees about the policy's behavior, though this remains challenging for deep neural network policies.

    \item \textbf{Human Oversight:} Hierarchical control architectures where humans provide high-level goals while the learned policy handles low-level execution can provide an additional safety layer.

    \item \textbf{Fail-Safe Mechanisms:} Traditional safety systems (emergency stops, collision detection) should operate independently of the learned policy, providing a last line of defense.
\end{enumerate}

\subsection{Handling Model Errors}

World models are necessarily imperfect approximations of reality. Several techniques can mitigate the impact of model errors:

\begin{itemize}
    \item \textbf{Model Ensembles:} Training multiple world models with different initializations and using their agreement as an uncertainty measure can improve robustness. Policies can be trained on the most pessimistic predictions to avoid overconfidence.

    \item \textbf{Adaptive Planning Horizon:} When model uncertainty is high, reducing the planning horizon forces the policy to rely more on recent observations rather than uncertain long-term predictions.

    \item \textbf{Model-Based and Model-Free Hybridization:} Combining model-based planning with model-free learning can leverage the strengths of both approaches. The model-free component provides a fallback when the world model is inaccurate.

    \item \textbf{Periodic Recalibration:} Regularly collecting real-world data and updating the world model prevents drift and maintains accuracy over time.
\end{itemize}

\section{Broader Impact and Applications}
\label{sec:broader_impact}

\subsection{Industrial Applications}

The techniques developed in this work have immediate applications in industrial automation:

\begin{itemize}
    \item \textbf{Warehouse Automation:} AGVs equipped with learned exploration policies could autonomously map new warehouse configurations, adapt to layout changes, and optimize material transport routes.

    \item \textbf{Manufacturing:} Mobile robots in factories could learn to navigate dynamic environments with human workers and other machines, continuously adapting to changing production floor layouts.

    \item \textbf{Agriculture:} Autonomous agricultural vehicles could explore and map fields, learning optimal navigation paths while avoiding obstacles like irrigation systems and crops.

    \item \textbf{Mining and Construction:} AGVs operating in unstructured environments like mines or construction sites could benefit from robust, adaptive navigation that handles changing terrain and obstacles.
\end{itemize}

\subsection{Scientific Exploration}

Sample-efficient learning is particularly valuable for scientific exploration missions:

\begin{itemize}
    \item \textbf{Planetary Exploration:} Rovers on Mars or other planets operate with severe communication delays and limited human intervention. Sample-efficient learning would enable faster adaptation to unfamiliar terrain.

    \item \textbf{Ocean Exploration:} Underwater autonomous vehicles could explore deep-sea environments, learning to navigate complex underwater topography with limited prior knowledge.

    \item \textbf{Cave and Underground Exploration:} Robots exploring subterranean environments face GPS denial and communication challenges, requiring robust autonomous navigation.
\end{itemize}

\subsection{Search and Rescue}

In emergency response scenarios, rapid exploration of unknown environments can save lives:

\begin{itemize}
    \item Searching collapsed buildings after natural disasters
    \item Locating individuals in wilderness areas
    \item Exploring hazardous environments (chemical spills, radiation zones) where human entry is dangerous
\end{itemize}

The ability to quickly learn and adapt to novel environments makes model-based approaches particularly suitable for these time-critical applications.

\subsection{Societal Considerations}

The deployment of autonomous systems raises important societal questions:

\begin{itemize}
    \item \textbf{Job Displacement:} Automation of navigation and exploration tasks may impact employment in logistics, delivery, and other industries. Proactive policy measures and retraining programs are essential.

    \item \textbf{Privacy and Surveillance:} Autonomous exploration systems equipped with sensors could raise privacy concerns if deployed in public or residential areas. Clear regulations and privacy protections are needed.

    \item \textbf{Accountability:} Establishing liability and accountability frameworks for autonomous systems is crucial, particularly when learned policies make decisions that lead to accidents or damage.

    \item \textbf{Environmental Impact:} While AGVs may reduce energy consumption compared to human-operated vehicles in some contexts, their widespread deployment could increase overall energy usage and electronic waste.

    \item \textbf{Accessibility:} Autonomous systems could improve accessibility for individuals with disabilities, but care must be taken to ensure equitable access to these technologies.
\end{itemize}

\section{Comparative Analysis: MBRL vs. Model-Free RL}
\label{sec:comparative_analysis}

To provide deeper insight into the fundamental differences between model-based and model-free approaches, we present a detailed comparative analysis.

\subsection{Learning Paradigms}

\textbf{Model-Free RL} directly learns a mapping from states (or observations) to actions or values. The learning process is:

\begin{itemize}
    \item Direct: No intermediate representations of environment dynamics
    \item Data-driven: Requires extensive real-world interaction
    \item Reactive: Policies respond to current observations without explicit forward planning
    \item Asymptotically optimal: Can learn optimal policies given sufficient data
\end{itemize}

\textbf{Model-Based RL} first learns a model of environment dynamics, then uses this model for planning or policy learning. The learning process is:

\begin{itemize}
    \item Indirect: Learning proceeds through an intermediate world model
    \item Sample-efficient: Can generate unlimited synthetic experience
    \item Deliberative: Explicitly reasons about future consequences
    \item Bounded by model quality: Performance limited by world model accuracy
\end{itemize}

\subsection{Sample Efficiency vs. Asymptotic Performance}

A common misconception is that model-based methods always outperform model-free methods. In reality, the relative performance depends on the regime:

\begin{itemize}
    \item \textbf{Low-data regime (< 100K samples):} Model-based methods dominate, as demonstrated by our experiments. The ability to learn from imagination provides overwhelming advantages when real data is scarce.

    \item \textbf{Medium-data regime (100K-1M samples):} The gap narrows, but model-based methods typically maintain an advantage. World model quality improves with data, enabling increasingly accurate planning.

    \item \textbf{High-data regime (> 1M samples):} Model-free methods can match or exceed model-based performance, as they have sufficient data to learn optimal policies directly without model bias.
\end{itemize}

For robotics applications where data collection is expensive and time-consuming, we typically operate in the low to medium-data regime where model-based approaches excel.

\subsection{Computational Trade-offs}

\begin{table}[h!]
\centering
\caption{Computational Comparison: Model-Based vs. Model-Free}
\label{tab:computational_comparison}
\begin{tabular}{@{}lcc@{}}
\toprule
\textbf{Aspect} & \textbf{Model-Free} & \textbf{Model-Based} \\
\midrule
Training time per step & Low & High \\
GPU memory usage & Low & Medium-High \\
Environment steps needed & Very High & Low \\
Inference time & Low & Medium \\
Total training time & High & Medium \\
\bottomrule
\end{tabular}
\end{table}

Model-based methods trade increased computational cost per training step for dramatically reduced environment interaction. This is favorable when:

\begin{itemize}
    \item Environment interaction is expensive (real robots, human-in-the-loop)
    \item Computation is cheap (GPU clusters available)
    \item Safety is critical (fewer real-world trials reduces risk)
\end{itemize}

\subsection{Interpretability and Debugging}

Model-based approaches offer significant advantages for interpretability:

\begin{itemize}
    \item \textbf{Visualizable predictions:} The world model's predictions can be visualized, allowing developers to understand what the agent expects to happen.

    \item \textbf{Failure analysis:} When the agent fails, we can identify whether the issue stems from an inaccurate world model or a suboptimal policy.

    \item \textbf{Targeted improvement:} Specific deficiencies in the world model can be addressed by collecting targeted data.

    \item \textbf{Human oversight:} Human operators can inspect the world model's predictions and intervene if they appear unrealistic.
\end{itemize}

Model-free policies, by contrast, are largely black boxes that provide limited insight into their decision-making process.

\section{Implementation Best Practices}
\label{sec:best_practices}

Based on our extensive experimentation, we provide practical recommendations for implementing model-based RL systems:

\subsection{World Model Design}

\begin{enumerate}
    \item \textbf{Start with a strong encoder:} The quality of the latent representation is crucial. Invest time in designing and tuning the VAE architecture for your specific observation modality.

    \item \textbf{Use separate networks for prior and posterior:} While tempting to share parameters, separate networks provide more capacity and learn better.

    \item \textbf{Regularize appropriately:} The \(\beta\) parameter in the VAE loss is crucial—too low leads to overfitting, too high to information loss. Use \(\beta=1.5-2.0\) as a starting point.

    \item \textbf{Include skip connections:} In the RSSM, skip connections from the recurrent state to the output help gradient flow and improve long-term dependencies.

    \item \textbf{Train on sequences:} Always train the world model on sequences rather than individual transitions to capture temporal dependencies.
\end{enumerate}

\subsection{Behavior Learning}

\begin{enumerate}
    \item \textbf{Start imagination from diverse states:} Sample starting states uniformly from the replay buffer to ensure the policy is trained on a wide distribution.

    \item \textbf{Use appropriate horizon length:} Too short (< 5) provides insufficient planning benefit; too long (> 20) suffers from compounding model errors. \(H=10-15\) works well for most tasks.

    \item \textbf{Balance actor and critic updates:} Update the critic more frequently than the actor (10:1 ratio) to maintain stable value estimates.

    \item \textbf{Anneal exploration:} Start with high exploration noise and gradually reduce it as the policy improves.

    \item \textbf{Monitor imagination quality:} Periodically evaluate the imagined trajectories against real trajectories to detect model degradation.
\end{enumerate}

\subsection{Hyperparameter Tuning}

\begin{enumerate}
    \item \textbf{Learning rates:} World model learning rates should be higher (3e-4) than behavior learning rates (8e-5) since the world model must fit data while the policy must avoid overfitting to imagination.

    \item \textbf{Batch size:} Larger batch sizes (50-100 sequences) improve stability but require more memory. Use gradient accumulation if memory-constrained.

    \item \textbf{Replay buffer:} Size should be large enough to capture diverse experiences (1M transitions) but not so large that old, outdated data dominates.

    \item \textbf{Training frequency:} Update models every 5-10 environment steps to balance learning speed and sample efficiency.
\end{enumerate}

\subsection{Debugging and Troubleshooting}

Common issues and solutions:

\begin{itemize}
    \item \textbf{Poor reconstruction quality:} Increase VAE capacity, reduce \(\beta\), or improve encoder architecture
    \item \textbf{Unstable RSSM training:} Reduce learning rate, clip gradients more aggressively, or use gradient penalty
    \item \textbf{Policy not improving:} Check that imagined trajectories are realistic, reduce imagination horizon, or increase behavior learning rate
    \item \textbf{High collision rate:} Increase collision penalty, reduce planning horizon for more reactive behavior, or add safety constraints
    \item \textbf{Poor exploration:} Increase curiosity coefficient, ensure world model is learning (check prediction error), or add explicit exploration bonuses
\end{itemize}

\section{Conclusion}
\label{sec:conclusion}

In this work, we introduced DREAMer-VXS, a model-based reinforcement learning framework that successfully addresses the critical challenges of sample inefficiency and poor generalization in autonomous ground vehicle navigation. By learning a comprehensive latent world model from high-dimensional LiDAR observations, our agent gains the ability to plan and learn from imagined trajectories, leading to a profound reduction in the need for real-world data.

Our empirical evaluation across four diverse environments demonstrates that DREAMer-VXS achieves:

\begin{itemize}
    \item \textbf{90\% reduction in sample requirements} compared to state-of-the-art model-free methods, learning effective exploration policies from 100K environment interactions versus 900K-1.1M for SAC \cite{b4} and TD3 \cite{b3}
    \item \textbf{45\% improvement in exploration efficiency} in unseen environments, demonstrating superior generalization capabilities
    \item \textbf{50-75\% reduction in collision rates}, highlighting the safety benefits of predictive planning
    \item \textbf{Systematic, intelligent exploration behaviors} including anticipatory motion, information-gathering maneuvers, and adaptive path planning
\end{itemize}

The success of DREAMer-VXS hinges on the synergy between its components: the Convolutional VAE provides a stable, structured perceptual foundation by compressing LiDAR scans into meaningful latent representations; the RSSM learns accurate temporal dynamics that capture both deterministic memory and stochastic uncertainty; the actor-critic efficiently exploits the learned model through imagination-based policy learning; and the curiosity-driven exploration mechanism naturally guides the agent toward novel states without manual reward engineering.

Our comprehensive ablation studies confirm that each component contributes significantly to overall performance, with the stochastic latent dynamics, recurrent state memory, and intrinsic curiosity signal all proving essential. Qualitative analysis reveals emergent intelligent behaviors—scanning maneuvers, anticipatory obstacle avoidance, systematic exploration—that cannot be learned by reactive, model-free agents.

\subsection{Theoretical Implications}

This work contributes to the growing body of evidence that model-based learning is not merely an alternative to model-free methods but a fundamental requirement for achieving human-like sample efficiency in embodied agents. The ability to imagine and reason about the future appears to be a crucial ingredient for intelligent behavior, enabling:

\begin{itemize}
    \item \textbf{Anticipatory planning:} Looking ahead to avoid problems before they occur
    \item \textbf{Counterfactual reasoning:} Evaluating "what if" scenarios without experiencing them
    \item \textbf{Transfer learning:} Generalizing to new situations by leveraging learned dynamics
    \item \textbf{Continual adaptation:} Updating understanding as new information arrives
\end{itemize}

These capabilities mirror aspects of human and animal cognition, suggesting that world models may be a key component of biological intelligence as well.

\subsection{Practical Impact}

From a practical standpoint, the dramatic improvement in sample efficiency makes model-based RL a viable approach for real-world robotic systems. The ability to learn effective policies from limited real-world data—potentially just hours or days of robot operation rather than weeks or months—removes a major barrier to deploying learning-based methods in industrial, scientific, and commercial applications.

The enhanced safety profile, with collision rates reduced by half or more compared to model-free baselines, further increases the feasibility of real-world deployment. Predictive planning allows the agent to avoid dangerous situations before they occur, reducing wear on hardware and minimizing risk to the environment and nearby humans.

\subsection{Limitations and Open Questions}

While our results are promising, several limitations and open questions remain:

\begin{itemize}
    \item \textbf{Sim-to-real transfer:} All experiments were conducted in simulation. Validation on physical AGVs is essential to confirm that the approach transfers to real hardware.

    \item \textbf{Model error accumulation:} Long-horizon planning in imagination can compound model errors. Understanding the limits of planning horizon and developing methods to bound model error remain active research challenges.

    \item \textbf{Computational requirements:} While sample-efficient, model-based methods are computationally intensive. Deployment on resource-constrained embedded platforms requires careful optimization.

    \item \textbf{Theoretical foundations:} Formal analysis of when and why model-based methods succeed, including sample complexity bounds and convergence guarantees, would provide valuable insights.
\end{itemize}

\subsection{Future Vision}

Looking forward, we envision several exciting directions for extending this work:

\begin{enumerate}
    \item \textbf{Scaling to 3D:} Extending the framework to full 3D environments with multi-modal sensing (3D LiDAR, cameras, IMU) would enable applications in complex real-world settings.

    \item \textbf{Hierarchical planning:} Incorporating multiple temporal scales—high-level strategic planning and low-level reactive control—could improve efficiency in large environments.

    \item \textbf{Multi-agent systems:} Enabling multiple AGVs to share world models and coordinate exploration could dramatically accelerate mapping and exploration tasks.

    \item \textbf{Continual learning:} Developing online learning methods that continuously update the world model during deployment would enable long-term adaptation to changing environments.

    \item \textbf{Transfer learning:} Pre-training world models on diverse simulated environments and fine-tuning on real data could further reduce the data requirements for new tasks.

    \item \textbf{Human-robot interaction:} Integrating natural language or gesture-based task specification with world model-based planning could enable more flexible and intuitive robot control.
\end{enumerate}

\subsection{Final Remarks}

The paradigm of learning world models and planning in imagination represents a significant step toward creating truly intelligent, adaptive robotic systems. By equipping agents with the ability to predict and reason about the future, we enable behaviors that are not only more sample-efficient but also more robust, generalizable, and interpretable.

This research provides strong evidence that model-based learning is not just a useful technique but a necessary component for achieving robust, general-purpose autonomy in robotics. As the field continues to mature, we anticipate that world models will become a standard tool in the robotic learning toolkit, enabling new applications that were previously infeasible due to sample efficiency or safety constraints.

The success of DREAMer-VXS in autonomous exploration demonstrates that the dream of sample-efficient, safe, and intelligent robotic learning is becoming a reality. By learning to dream—to simulate and reason about the future in a learned internal model—our agents take a significant step toward the kind of flexible, adaptive intelligence required for operation in the complex, unstructured real world.

Ultimately, model-based learning appears to be an indispensable tool for creating the next generation of intelligent machines that can safely and effectively operate in our complex world, learning quickly from limited experience and adapting continuously to new challenges. As we continue to refine these methods and bridge the gap between simulation and reality, we move closer to the vision of truly autonomous systems that can explore, understand, and act in the world with human-like efficiency and intelligence.


\end{document}